\documentclass[manuscript,screen,11pt]{acmart}

\renewcommand\footnotetextcopyrightpermission[1]{}
\usepackage{booktabs}
\usepackage[table]{xcolor}
\usepackage{makecell}
\usepackage{tikz}
\usepackage{url}

\usetikzlibrary{shapes.arrows,positioning,matrix,backgrounds,arrows.meta}

\definecolor{wideColor}{HTML}{1B5E20}
\definecolor{longcpColor}{HTML}{8BCAD5}
\definecolor{blueA}{RGB}{33,114,255}
\definecolor{redA}{RGB}{220,53,69}
\definecolor{grayA}{RGB}{140,140,140}

\tikzset{
  dagNodeLinear/.style = {
      circle,
      minimum size=7.5mm,
      inner sep=0pt,
      font=\footnotesize\bfseries,
      draw=blue!40!black,
      very thick,
      fill=blue!35
  },
  dagNodeGNP/.style = {
      circle,
      minimum size=7.5mm,
      inner sep=0pt,
      font=\footnotesize\bfseries,
      draw=red!55!black,
      very thick,
      fill=red!65!black
  },
  dagEdgeLinear/.style = {
      ->,
      line width=1.1pt,
      draw=black!70,
      >=Latex
  },
  dagEdgeGNP/.style = {
      ->,
      line width=1.0pt,
      draw=gray!70,
      >=Latex
  },
  dagHalo/.style = {
      fill=blue!10,
      draw=none,
      rounded corners=2pt
  },
  dagTitle/.style = {
      font=\bfseries\small,
      text=black!75
  }
}

\title{
On the Role of DAG topology in Energy-Aware Cloud Scheduling : A GNN-Based Deep Reinforcement Learning Approach
}

\author{Anas Hattay}
\email{anas.hattay@cea.fr}
\affiliation{
  \institution{CEA, List, Université Paris-Saclay}
  \city{Palaiseau}
  \country{France}
}

\author{Fred Ngole Mboula}
\email{fred-maurice.ngole-mboula@cea.fr}
\affiliation{
  \institution{CEA, List, Université Paris-Saclay}
  \city{Palaiseau}
  \country{France}
}

\author{Eric Gascard}
\email{eric.gascard@grenoble-inp.fr}
\affiliation{
  \institution{Université Grenoble Alpes, CNRS, Grenoble INP, G-SCOP}
  \city{Grenoble}
  \country{France}
}

\author{Zakaria Yahouni}
\email{zakaria.yahouni@grenoble-inp.fr}
\affiliation{
  \institution{Université Grenoble Alpes, CNRS, Grenoble INP, G-SCOP}
  \city{Grenoble}
  \country{France}
}

\begin{document}


\begin{abstract}
Cloud providers must assign heterogeneous compute resources to workflow DAGs while balancing competing objectives such as completion time, cost, and energy consumption. In this work, we study a single-workflow, queue-free scheduling setting and consider a graph neural network (GNN)–based deep reinforcement learning scheduler designed to minimize workflow completion time and energy usage.

We identify specific out-of-distribution (OOD) conditions under which GNN-based deep reinforcement learning schedulers fail, and provide a principled explanation of why these failures occur. Through controlled OOD evaluations, we demonstrate that performance degradation stems from structural mismatches between training and deployment environments, which disrupt message passing and undermine policy generalization. Our analysis exposes fundamental limitations of current GNN-based schedulers and highlights the need for more robust representations to ensure reliable scheduling performance under distribution shifts.
\end{abstract}



\maketitle
\pagestyle{empty}

\section{Introduction}
\label{sec:introduction}

\subsection{Context}

Over the past few years, artificial intelligence and large language models have pushed cloud systems much harder than before. According to \cite{IEA2025DataCenters}, the energy used by data centers keeps growing because of heavier AI training and inference workloads. 
This means even small improvements in how cloud resources are allocated can save a noticeable amount of time and electricity.

\begin{figure*}[htp]
    \centering
    \includegraphics[width=\linewidth]{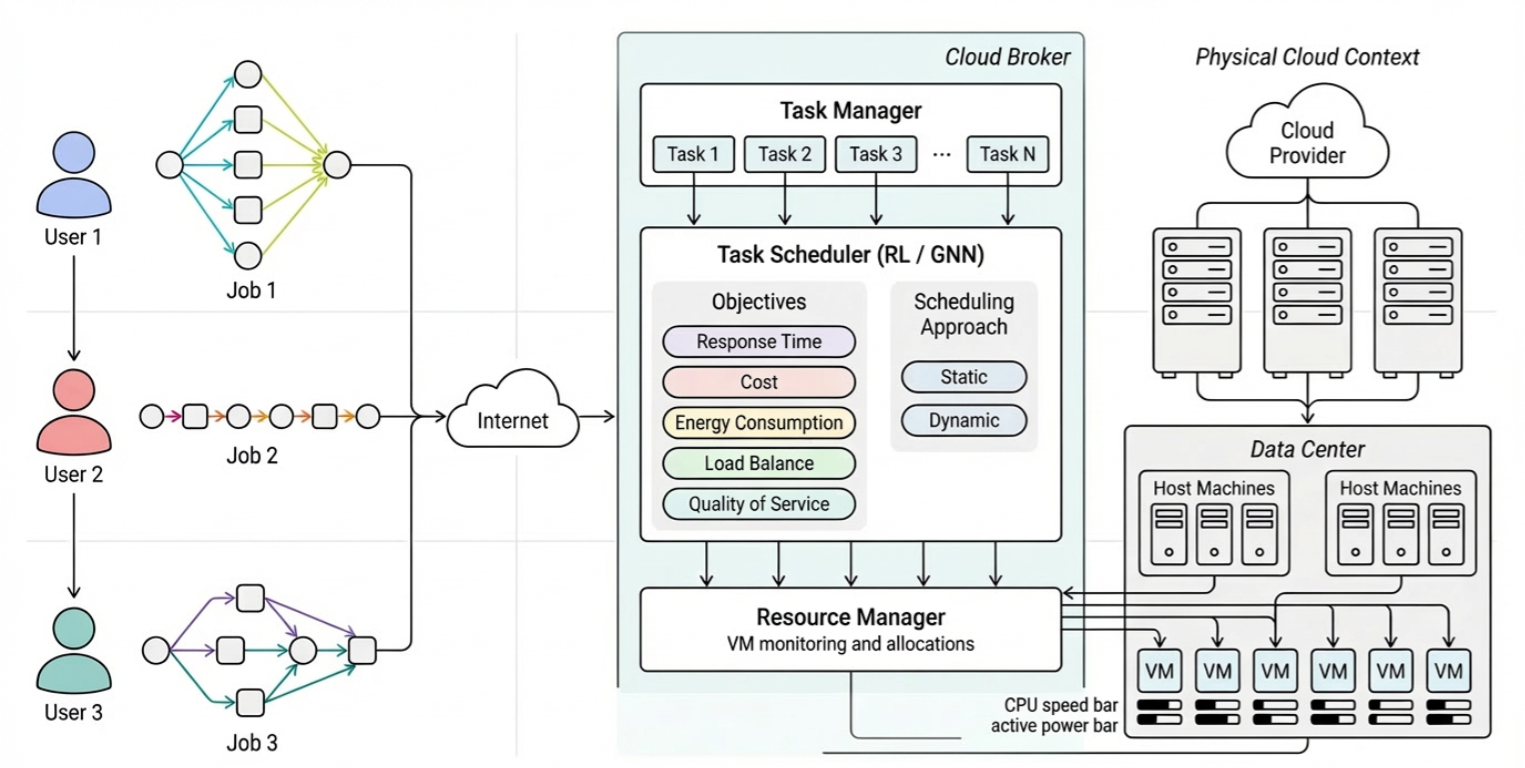}
\caption{\textbf{Overview of the workflow scheduling problem in a heterogeneous cloud}. The scheduler receives a DAG with ready tasks and a set of machines with different speed and power. It must decide which task runs on which machine, and each choice changes both completion time and energy.}
    \label{fig:Pb definition}
\end{figure*}

Figure~\ref{fig:Pb definition} shows the setting. A workflow is a directed acyclic graph. Nodes are tasks. Edges are data or control dependencies. Some stages expose wide parallelism. Others form long serial chains. At the same time, machines in a cloud are heterogeneous. Some machines finish tasks faster but draw more power. Others are slower but use less. The scheduler decides, at each step, which ready task should run on which machine. Each assignment changes the makespan and the total energy used.

This problem has drawn steady research interest. Energy-aware scheduling appeared in HPC and cloud literature well before the recent AI boom. Early work used DVFS to trade performance for power \citep{Zong2007DVFS, Rountree2009DVFS}. Recent surveys report steady activity: Multiple systematic reviews published indicate sustained research activity in this domain \citep{Versluis2020WorkflowSurvey, Ajmera2024VMScheduling}. Machine learning-based cluster management has demonstrated measurable impact, reducing data center energy consumption by up to 15\% in Google's production deployments \citep{GoogleDataCenterAI}. Amazon and Microsoft have explored complementary approaches through right-sizing and spot market mechanisms to reduce both operational costs and resource waste \citep{Cortez2017AWS, Shahrad2020Serverless}.

Modern AI workflows add new dimensions to this problem. They mix layers that are very wide with chains that are very long. And they run on machines with big differences in speed and power use. Old scheduling rules assume the workflow shape stays the same and machines are predictable. When that is not true, performance can drop fast. Because of this, researchers are looking at more advanced schedulers that can adapt. But the problem is we do not know if these learned policies still work when the workflow or machines change. We will go over the current approaches next and then introduce the gap that motivated our study.

\subsection{Related Work}

Many workflow applications are naturally written as directed acyclic graphs (DAGs) of tasks with different resource needs and complex dependencies (\cite{deelman2009workflows}, \cite{sakellariou2010mapping}). Scheduling these DAGs on pools of virtual machines with different speeds and power use is a classic problem in distributed systems and high performance computing (\cite{mao2012survey}). This problem is NP hard in general. Foundational work by \citet{Pinedo2012} and the notation of \citet{Graham1979} provide the standard formal basis for makespan minimization under precedence and machine heterogeneity.

\subsubsection{Classical Topology Aware DAG Scheduling}

A large body of work models applications as weighted DAGs, where nodes are tasks and edges capture precedence and communication constraints on heterogeneous processors. The Heterogeneous Earliest Finish Time (HEFT) and Critical Path on a Processor (CPOP) list schedulers of \citet{Topcuoglu2002HEFT} are central references. They rank tasks with path based metrics and map them to heterogeneous processors to reduce makespan. HEFT uses an upward rank that approximates remaining critical path length, then schedules tasks on the processor with the earliest finish time. This exploits DAG depth, fan out, and communication along paths. CPOP identifies a global critical path and assigns its tasks to a single processor to reduce inter processor communication.

Many variants build on this topology aware idea while adding new objectives such as energy or reliability. Energy aware DAG schedulers on heterogeneous and embedded platforms often use critical path and level based slack to decide which tasks can safely be slowed down or consolidated while still meeting deadlines \citep{hu2023online}. In these works, DAG topology is not only a feasibility constraint. It is a direct signal for making decisions. Parallelism per level, critical path length, and slack structure strongly influence task priorities and mapping choices.

In cloud and distributed environments, workflow systems such as Pegasus \citep{deelman2015pegasus} and simulators such as CloudSim \citep{Calheiros2011CloudSim} have established DAGs as the standard abstraction for scientific workflows. Energy aware scheduling has been widely studied in this setting. For example, \citet{Beloglazov2012Energy} show how power heterogeneity and job topology together shape good placement policies.

Many real world workloads are also multi objective. Schedulers must trade off makespan, cost, energy, and sometimes reliability (\cite{bittencourt2018scheduling}, \cite{dasilva2017learning}). Classical heuristics are usually tuned around a single dominant objective. When new metrics are added, they often need extensive manual retuning and may not transfer well across workloads \citep{bittencourt2018scheduling}. Meta heuristic methods can explore richer trade offs but are often too slow for real time scheduling in dynamic clouds, since they rely on iterative search (\cite{dasilva2017learning}).

\subsubsection{Deep Reinforcement Learning for Resource Scheduling}

These limitations have motivated a shift toward learning based schedulers. Deep reinforcement learning (deep RL ), first introduced by \citet{sutton1998reinforcement}, offers a way to learn scheduling policies directly from experience without manually encoding rules for every scenario. Early work by \citet{mao2016resource} showed that deep RL  agents could learn to pack tasks and allocate resources in cluster settings, outperforming hand tuned heuristics on job completion time. The idea is to frame scheduling as a sequential decision problem. An agent observes system state, selects actions such as which task to schedule or which machine to use, and receives rewards based on performance metrics like makespan or resource use.

This approach has several advantages. First, the policy can adapt to patterns in the workload without explicit feature engineering. Second, it can handle multi objective trade offs by shaping the reward function. Third, once trained, the policy can make decisions quickly, which is useful in online settings. Several studies have confirmed that deep RL  schedulers can match or beat classical heuristics in controlled environments (\cite{zhang2021deepjs}, \cite{mao2016resource}).

But early deep RL  schedulers often treated system state as a flat feature vector. This does not capture the relational structure of workflows or resource topologies. This is where graph neural networks come in.

\subsubsection{Graph Neural Networks for Structured Scheduling}

Graph neural networks provide a natural way to encode relational structure. In scheduling, both workflows (task dependencies) and resource topologies (machine connectivity or hierarchy) are naturally represented as graphs. GNNs use message passing to let each node aggregate information from its neighbors. This means the learned representation can respect the structure of the problem.

Decima was one of the first systems to combine GNNs with RL for cluster scheduling \citep{Mao2019Decima}. It models each data processing job as a DAG of stages. A GNN embeds this DAG along with per stage features such as remaining work and resource demand. The RL policy then uses these embeddings to decide which stage to schedule and how many executors to assign. Message passing over the DAG lets the model implicitly capture properties like depth, critical paths, and fan in or fan out. Experiments showed that this topology aware policy reduced average job completion time compared to topology agnostic baselines and classical heuristics.

Follow up work extended this idea to other scheduling domains. \citet{park2021learning} used GNNs to embed both job DAGs and cluster topology for multi resource scheduling. Others applied similar architectures to workflow scheduling in cloud and edge environments, where tasks have precedence constraints and machines have different speeds or power profiles. In most cases, the GNN based approach improved performance over flat feature representations. These results suggest that exploiting the graph structure provides richer relational information, allowing the policy to generalize more robustly across scenarios.

\subsubsection{Generalization and Robustness Challenges}

Despite these successes, most studies evaluate learned policies primarily on the same types of workflows and host configurations seen during training. They show that GNN based RL can work well in specific settings, but they rarely ask how robust these policies are when the workflow structure or resource characteristics change. For example, what happens when a policy trained on shallow parallel workflows is tested on deep sequential workflows? Or when a policy trained on homogeneous machines is deployed on heterogeneous hardware with conflicting speed and power trade offs?

This concern is grounded in known limitations of graph neural networks under distribution shift.  \citet{wu2022handling} showed at ICLR 2022 that distribution shifts hit GNNs particularly hard because of how nodes connect to each other. When the graph topology changes, the whole representation can break down. This matters for workflow scheduling. A policy trained on one type of DAG may face very different graphs at deployment. Depth can change. Width can change. Branching can change. This motivates a systematic first step: characterize the distribution shift in our setting. We must precisely define and measure how deployment DAGs differ from those seen during training, and then determine how to address it.

\subsection{Positioning of Our Work}

This paper identifies specific out-of-distribution conditions that cause GNN-based deep RL schedulers to fail, and explains why these failures occur.

We build on two key observations. First, \citet{9460684} showed that real workflows from production systems (millions of DAGs from Alibaba batch jobs) cluster into a few structural types. Second, recent RL-based schedulers using GNNs to embed DAGs have shown strong performance on some benchmarks (\cite{Mao2019Decima}, \cite{park2021learning}), but their robustness across structural types remains unclear.

An empirical pattern is reported in \cite{hattay2024evaluating}. When a deep RL scheduler is compared with classical heuristics across many workflow instances, it does not simply dominate or fail everywhere. It performs very well on some workflows and host settings and quite poorly on others. Sometimes it is even worse than basic list scheduling rules. With closer examination, we observed that these failures are not random. They concentrate in specific combinations of workflow topology and host heterogeneity. This suggests that learned schedulers have implicit structural domains where they behave coherently and domains where their behavior degrades.

Epistemologically, we follow a Popper style view of scientific progress \citep{popper2005logic}. We are less interested in showing more positive cases for a learned policy and more interested in subjecting it to tests that might break it. We treat the learned policy as a provisional theory about how topology and heterogeneity shape scheduling decisions. We then expose that theory to workflow topologies and host regimes in which it should fail if it is narrow or brittle. The goal is not only to show that the agent can work somewhere, but to reveal what it has actually learned when the surrounding conditions change.

To study this in a controlled way, we define two simple workflow families. \emph{wide} DAGs are shallow with high parallelism. \emph{Long Critical Path (LongCP)} DAGs have deep dependency chains and little parallel slack. On the resource side, we consider four queue free host regimes that isolate different aspects of heterogeneity: Homogeneous Speed (HS), Homogeneous Power (HP), Heterogeneous Aligned (AL), and Heterogeneous Non Aligned (NA). Each regime creates a different trade off between makespan and active energy. As a result, each regime gives a different incentive for using or ignoring parallelism.

We then train a GNN based actor critic deep RL scheduler on these environments. We use separate agents specialized to wide workflows and to LongCP workflows. We do not evaluate these agents only on the distributions they were trained on. Instead, we systematically probe cross topology and cross regime generalization. For example, we run a wide trained agent on LongCP workflows, a LongCP trained agent on wide workflows, and we test all agents across all four host regimes.

This work therefore investigates how DAG topology and host heterogeneity together shape the behavior and generalization of an RL based scheduler with joint energy and makespan objectives in a queue free, single workflow cloud setting. 

\paragraph{Research Questions}

This setup lets us ask two main questions:

\begin{itemize}
    \item \textbf{Q1:} \label{q:joint-influence} How do DAG topology (wide vs. long critical path) and host speed/power configurations jointly influence learned policy priorities under mixed energy-makespan objectives? Do these factors induce systematic biases in scheduling strategies despite identical objectives?

    \item \textbf{Q2:} \label{q:cross-generalization} How does cross-topology generalization vary across host configurations (AL, NA, HS, HP)? When do wide specialists outperform LongCP specialists (and vice versa), and what explains these performance gaps?

\end{itemize}

\paragraph{Contributions}

By answering these questions, we make the following contributions:

\begin{itemize}
  \item \textbf{A controlled decomposition of the problem space.} We separate the effects of workflow topology and host heterogeneity by defining two contrasting DAG families (wide and LongCP) and four host regimes (HS, HP, AL, NA) that each capture a different dimension of heterogeneity. 
  

  \item \textbf{Systematic cross topology and cross regime evaluation.} We train GNN based deep RL schedulers specialized to wide workflow and to LongCP workflows. We then test all agents across all topology and regime combinations using wide and LongCP test workflows per configuration.

  \item \textbf{An interpretability focused analysis of generalization domains.} We show that the combined structure of workflow and host naturally divides the problem into domains where a given policy behaves consistently and domains where its generalization breaks down.

\end{itemize}

In the following sections, we first formalize the problem and objectives (Section~\ref{sec:Pb}), then describe our benchmark and GNN based scheduler (Section~\ref{sec:architecture}), and finally evaluate and explain its behavior across workflow topologies and host regimes (Section~\ref{sec:exp_methodology}).

\section{Problem Setup}
\label{sec:Pb}
\subsection{Problem Formulation}
\label{sec:PF}

We adopt the Markov Decision Process (MDP) formulation from \citet{chandrasiri2025energy} for workflow scheduling on virtualized clusters, with modifications to support concurrent task execution on multi-core VMs. While we retain the same state space, action space, and transition dynamics, we extend the makespan and active energy calculations to account for overlapping tasks running on a single VM and fractional CPU utilization when computing energy consumption and VM resource liberation times.

We model workflow scheduling as a finite-horizon MDP
\[
\mathcal{MDP}=(\mathcal{S},\mathcal{A},\mathcal{P},\mathcal{R},\gamma),
\]
with state space $\mathcal{S}$, action space $\mathcal{A}$, transition kernel $\mathcal{P}$, reward function $\mathcal{R}$, and discount factor $\gamma$. Decisions occur at epochs $k=0,\dots,K$ until all tasks complete.

\subsubsection{System Model.}
Let $\mathcal{G}=(\mathcal{V},\mathcal{E})$ be a workflow DAG with tasks $i\in\mathcal{V}$ and precedence edges
$(p\!\to\!i)\in\mathcal{E}$. Each task has:
(i) computational size $L_i$, representing the total amount of processing required to complete the task (e.g., MI),
(ii) resource demand vector $\mathbf{d}_i=(\mathrm{cpu}_i,\mathrm{mem}_i)$,
(iii) compatibility set $\mathcal{C}_i\subseteq\mathcal{M}$ of admissible VMs.
Each VM $m\in\mathcal{M}$ has capacity $\mathbf{c}_m=(C^{\mathrm{cpu}}_m,C^{\mathrm{mem}}_m)$,
processing speed $s_m$ (e.g., MIPS), and power parameters $(P^{\mathrm{idle}}_m,P^{\mathrm{peak}}_m)$.

\subsubsection{Decision Epochs and Clock.}
Let $\tau_k$ denote the simulation clock at decision epoch $k$. Decisions occur when the agent assigns a ready task.
Between decisions, the environment advances $\tau$ according to scheduled start/finish events implied by previous assignments.

\subsubsection{State Space $\mathcal{S}$.}
A state $s_k \in \mathcal{S}$ summarizes all information needed for optimal control:
\begin{align*}
s_k = (&\tau_k; 
\{\text{task status}_i \in \{\text{not\_ready},\text{ready},\text{running},\text{done}\}\}_{i\in\mathcal{V}};\\
&\{\text{parent\_ready}_i = \max_{p\in \mathrm{Pa}(i)} c_p \}_{i\in\mathcal{V}};\\
&\{\text{assignment}_i \in \mathcal{C}_i \cup \{\varnothing\}\}_{i\in\mathcal{V}};\\
&\{\text{start}_i,\,c_i\}_{i\in\mathcal{V}};\\
&\{\text{VM residual capacities and active allocations over }[\tau_k,\infty)\}_{m\in\mathcal{M}};\\
&\{(P^{\mathrm{idle}}_m, P^{\mathrm{peak}}_m)\}_{m\in\mathcal{M}};\\
&\{\mathcal{C}_i\}_{i\in\mathcal{V}} ).\end{align*}
\begin{itemize}
    \item $\mathrm{Pa}(i)$ denotes the set of parent tasks of task $i$ in the workflow DAG.
    \item $c_p$ is the completion time of parent task $p$.
    \item $\text{start}_i$ and $c_i$ are the planned start and completion times of task $i$.
    \item VM residual capacities and active allocations are tracked for each VM $m \in \mathcal{M}$ over future time.
    \item $(P^{\mathrm{idle}}_m, P^{\mathrm{peak}}_m)$ are the power parameters of VM $m$.
    \item $\mathcal{C}_i \subseteq \mathcal{M}$ is the set of VMs compatible with task $i$.
\end{itemize}

The ready set at $\tau_k$ is $\mathcal{R}_k=\{i:\ \text{task status}_i=\text{ready}\}$.

\subsubsection{Action Space $\mathcal{A}$.}
An action selects a task--VM pair:
\[
a_k=(i,m)\in \mathcal{F}(s_k)\subseteq \mathcal{V}\times\mathcal{M},
\]
where the feasible set enforces precedence, compatibility, and capacity:
\[
\mathcal{F}(s_k)=\Bigl\{(i,m):\ i\in\mathcal{R}_k,\ m\in\mathcal{C}_i,\ 
\mathbf{d}_i \preceq \text{residual\_capacity}_m(t)\ \text{for some } t\ge \text{parent\_ready}_i \Bigr\}.
\]
Operationally, assigning $(i,m)$ schedules task $i$ on VM $m$ at its earliest feasible start time
\[
s_i = \min \{ t \ge \text{parent\_ready}_i : \mathbf{d}_i \preceq \text{residual\_capacity}_m(t) \},
\]
and updates VM $m$'s capacity timeline, with completion time
\[
c_i = s_i + \frac{L_i}{s_m}, \quad \text{where } s_m \text{ is the processing speed of VM } m.
\].

 In contrast to \cite{chandrasiri2025energy}, which assumes one task per VM at a time, our formulation allows multiple tasks to execute concurrently on VM $m$ as long as the aggregate resource demands satisfy $\sum_{j\in A_m(t)} \text{cpu}_j \le C^{\text{cpu}}_m$ and $\sum_{j\in A_m(t)} \text{mem}_j \le C^{\text{mem}}_m$ for all $t$. Action masking enforces $\mathcal{F}(s_k)$.

\subsubsection{Transition Kernel $\mathcal{P}$.}
Given a state $s_k$ and an action $a_k = (i,m)$, the environment deterministically transitions to $s_{k+1}$ by performing the following updates:
(i) task $i$'s $(\text{start}_i,c_i,\text{status}_i)$,
(ii) VM $m$'s allocation timeline,
(iii) descendants' readiness when all parents are completed:
$\text{task status}_j\gets\text{ready}$ if $\forall p\in \mathrm{Pa}(j): \text{task status}_p=\text{done}$ and $\tau\ge \max_{p\in\mathrm{Pa}(j)} c_p$,
(iv) simulation clock to the next decision epoch $\tau_{k+1}$, i.e., the earliest time when a task becomes ready.

\subsubsection{Reward $\mathcal{R}$ with Concurrency-Aware Heuristics.}
\label{par:reward}
Designing informative intermediate rewards for workflow scheduling is challenging due to delayed objectives : makespan and total energy consumption. To enable effective credit assignment during learning, we define per-step rewards based on regret reductions relative to concurrency-aware heuristic estimates that approximate the remaining cost-to-go.

\subparagraph{\textbf{Heuristic Estimates.}}
At each state $s$, we compute two greedy estimates:
\begin{itemize}
    \item \textbf{Makespan estimate} $\widehat{T}(s)$: Earliest completion time obtained by greedily scheduling remaining tasks using an earliest-completion-time (ECT) policy.
    \item \textbf{Active energy estimate} $\widehat{E}(s)$: Minimum active energy consumption for remaining tasks, accounting for fractional CPU utilization and concurrent execution.
\end{itemize}

Both heuristics simulate a feasible completion of the workflow by:
\begin{enumerate}
    \item Building per-VM event timelines from already-scheduled tasks, where each event $(t, \Delta_{\text{mem}}, \Delta_{\text{cores}})$ tracks resource changes at time $t$.
    \item For each unscheduled task $i$, finding the earliest feasible start time $t\ge t_{\text{ready}}^i$ on each compatible VM $m\in\mathcal{C}_i$ when:
    \[
    \text{used\_mem}_m(t) + \text{mem}_i \le C^{\text{mem}}_m \quad\text{and}\quad \text{used\_cores}_m(t) + \text{cpu}_i \le C^{\text{cpu}}_m
    \]
    \item Selecting the VM that minimizes completion time (for makespan) or energy consumption (for energy).
\end{enumerate}

For the energy heuristic, power on VM $m$ at time $t$ is modeled as:
\[
P_m(t) = P^{\text{idle}}_m + (P^{\text{peak}}_m - P^{\text{idle}}_m) \cdot U_m(t),
\quad
U_m(t) = \min\!\left(1,\ \frac{1}{C^{\text{cpu}}_m}\sum_{j\in A_m(t)} \text{cpu}_j\right)
\]
where $A_m(t)$ is the set of tasks active on VM $m$ at time $t$, and $U_m(t)\in[0,1]$ is the fractional CPU utilization. This fractional CPU model extends \citet{chandrasiri2025energy} work by accounting for the aggregate core usage of all concurrent tasks, enabling accurate energy estimation when multiple tasks overlap on multi-core VMs. Energy is integrated piecewise-constant over segments bounded by task start/completion events.

\subparagraph{\textbf{Regret-Based Reward.}}
At each decision epoch $k$, after action $a_k$ transitions $s_k\to s_{k+1}$, we compute normalized regret reductions:
\[
\Delta R^{\text{mk}}_k = -\frac{\widehat{T}(s_{k+1}) - \widehat{T}(s_k)}{\max(\widehat{T}(s_{k+1}), \varepsilon)},
\qquad
\Delta R^{\text{en}}_k = -\frac{\widehat{E}(s_{k+1}) - \widehat{E}(s_k)}{\max(\widehat{E}(s_{k+1}), \varepsilon)}
\]
where $\varepsilon>0$ is a small constant. A positive value indicates the action reduced the estimated cost-to-go. The combined reward is:
\[
r_k = w_T \cdot \Delta R^{\text{mk}}_k + w_E \cdot \Delta R^{\text{en}}_k
\]
where $(w_T, w_E)$ are tunable weights that scalarize the multi-objective problem.

\subparagraph{\textbf{Objective.}}
We optimize a stationary policy $\pi_\theta(a\mid s)$ to maximize the expected discounted return
\[
J(\pi_\theta) = \mathbb{E}_{\pi_\theta,\mathcal{P}}\!\left[\sum_{k=0}^{K} \gamma^k\, r_k\right],
\]
where $\mathcal{P}$ denotes the environment transition dynamics, $r_k$ is the reward at decision step $k$, and $K$ is the episode horizon corresponding to the total number of scheduling decisions until termination. We typically set $\gamma \approx 1$ for episodic scheduling problems.

At episode termination, the final makespan $T_{\mathrm{mk}} = \max_i c_i$ and the total energy consumption
\[
E_{\mathrm{tot}} = \sum_{m\in\mathcal{M}} \int_{0}^{T_{\mathrm{mk}}} P_m(t)\,dt
\]
are computed for evaluation, where $P_m(t)$ denotes the instantaneous power consumption of machine $m$ at time $t$.

\subsubsection{Constraints and Termination.}
Feasibility is enforced by $\mathcal{F}(s_k)$:
(i) precedence constraints via readiness,
(ii) per-VM resource capacities (CPU cores and memory) over time with concurrent task support,
(iii) task--VM compatibility.
The episode terminates when all tasks are completed.

\subsection{Workflow Topology and Host Regime Decomposition}
\label{sec:mostimportant}

To understand the problem complexity, we first examine what makes one job different from another at a structural level. Since jobs consist of multiple interdependent tasks, their \emph{dependency structure} determines how many tasks can be ready in parallel and how scheduling decisions propagate through time.

Let $G=(V,E)$ be a DAG with task work $\{L_i\}_{i\in V}$. We denote:
\begin{itemize}
  \item total work $W = \sum_{i\in V} L_i$,
  \item critical-path length
    $L_{\mathrm{CP}} = \max_{\pi\in\text{paths}(G)} \sum_{i\in\pi} L_i$,
  \item depth $D$ and level widths $|\text{level}(\ell)|$ from a standard
        levelization of the DAG.
\end{itemize}
A useful scalar summary of intrinsic parallelism is
\[
  \Phi \;=\; \frac{W}{L_{\mathrm{CP}}}.
\]
Intuitively, $L_{\mathrm{CP}}$ is the amount of work that \emph{must} be done
sequentially, while $W$ is the total work. Large
$\Phi$ indicates ample exploitable parallelism; $\Phi$ close to $1$ indicates
a nearly sequential job.

We focus on two representative topologies:
\begin{itemize}
\item \textbf{Long Critical Path (LongCP).} Deep dependency chains ($D$ large)
  with small width, so $\Phi$ is close to $1$--few. Ready sets are typically
  small and concentrated near occasional side branches.
\item \textbf{wide DAG.} Shallow depth ($D$ small) with large level widths,
  so $\Phi \gg 1$. Ready sets are large and bursty at wide layers, and many
  task--VM assignments are simultaneously feasible. 
\end{itemize}

\begin{figure}[htp]
  \centering
  \begin{tikzpicture}[
      node/.style={circle,draw=wideColor,thick,fill=white,inner sep=1.8pt},
      edge/.style={-Latex,thin,draw=wideColor},
      >=Latex
    ]
    \node[node] (s) at (0,1.5) {};

    \node[node] (a1) at (-1,0.5) {};
    \node[node] (a2) at ( 0,0.5) {};
    \node[node] (a3) at ( 1,0.5) {};

    \node[node] (b1) at (-1.5,-0.5) {};
    \node[node] (b2) at (-0.5,-0.5) {};
    \node[node] (b3) at ( 0.5,-0.5) {};
    \node[node] (b4) at ( 1.5,-0.5) {};
    \node[node] (b5) at ( 0,-1.0)   {};

    \node[node] (t) at (0,-1.8) {};

    \foreach \x in {a1,a2,a3} \draw[edge] (s) -- (\x);

    \foreach \x in {a1,a2,a3}{
      \foreach \y in {b1,b2,b3,b4,b5}{
        \draw[edge,opacity=0.6] (\x) -- (\y);
      }
    }

    \foreach \x in {b1,b2,b3,b4,b5} \draw[edge] (\x) -- (t);

    \node[above=1mm of s] {\small \textbf{wide DAG}};
  \end{tikzpicture}
  \hspace{1.6em}
  \begin{tikzpicture}[
      node/.style={circle,draw=longcpColor,thick,fill=white,inner sep=1.8pt},
      edge/.style={-Latex,thin,draw=longcpColor},
      >=Latex
    ]

    \node[node] (c0) at (-1.2,1.5) {};
    \node[node] (c1) at (-0.4,1.0) {};
    \node[node] (c2) at ( 0.4,0.5) {};
    \node[node] (c3) at ( 1.2,0.0) {};
    \node[node] (c4) at ( 0.4,-0.5) {};
    \node[node] (c5) at (-0.4,-1.0) {};
    \node[node] (c6) at (-1.2,-1.5) {};
    \node[node] (c7) at ( 0.0,-1.8) {};

    \node[node] (b2a) at ( 1.1,0.9)  {};
    \node[node] (b3a) at ( 1.8,0.1)  {};
    \node[node] (b4a) at ( 1.1,-0.6) {};
    \node[node] (b5a) at (-1.8,-0.9) {};

    \foreach \u/\v in {c0/c1,c1/c2,c2/c3,c3/c4,c4/c5,c5/c6,c6/c7}{
      \draw[edge] (\u) -- (\v);
    }

    \draw[edge] (c2) -- (b2a);
    \draw[edge] (c3) -- (b3a);
    \draw[edge] (c4) -- (b4a);
    \draw[edge] (c5) -- (b5a);

    \node[above=1mm of c0] {\small \textbf{Long Critical Path DAG}};
  \end{tikzpicture}

  \caption{Schematic comparison of a
  \textcolor{wideColor}{\textbf{wide}} DAG (left, shallow with many parallel
  branches) and a \textcolor{longcpColor}{\textbf{Long‑CP}} DAG (right, deep
  dependency chain with limited side‑branch concurrency).}
  \label{fig:wide_vs_longcp_tikz}
\end{figure}
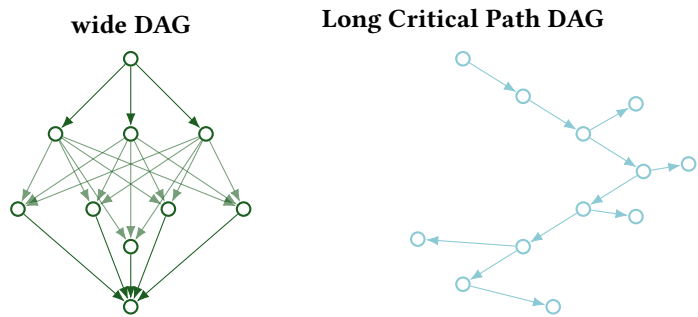

\begin{figure}[htp]
    \centering
    \includegraphics[width=0.75\textwidth]{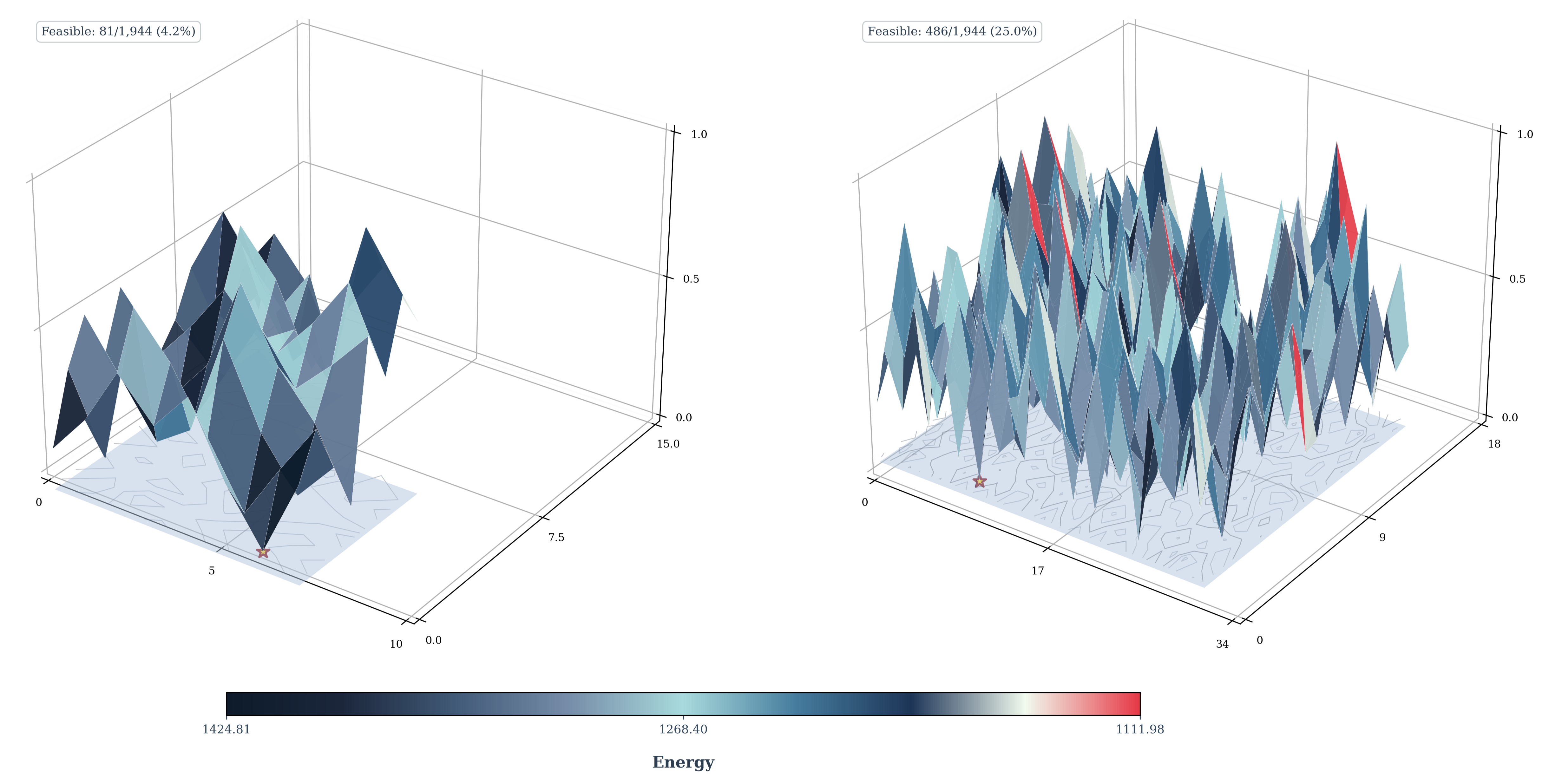}
    \caption{
        \textbf{Fitness landscape of the scheduling search space.}
        We exhaustively enumerate all feasible action sequences for a small scenario and project them onto a 2D plane using a Hilbert curve. The Z-axis shows energy consumption (lower is better). Left: LongCP DAG. Right: wide DAG .
    }
    \Description{A pair of side‑by‑side 3D landscapes compares scheduling fitness for two small task graphs. In each panel, every feasible scheduling action sequence is plotted as a point on a 2D plane generated by a Hilbert space‑filling curve; nearby coordinates represent similar sequences, and axis units are indices without physical meaning. The vertical axis is energy consumption, with lower height indicating a more efficient schedule.

The surface forms peaks, valleys, and plateaus. Valleys and basins mark clusters of low‑energy schedules. Peaks and ridges indicate schedules or neighborhoods with higher energy. Sharp drops or cliffs reflect cases where small changes in the action sequence cause large changes in energy, while smooth regions indicate insensitivity to local perturbations.

The left panel corresponds to a graph with a long critical path; the right panel corresponds to a graph with many parallel branches. The two surfaces differ in the location, size, and apparent density of low‑energy regions, illustrating how graph topology affects the distribution of good schedules and the ruggedness of the search space. Readers can use these landscapes to judge whether efficient schedules are concentrated in a few basins or dispersed across many regions, and whether exploring locally similar schedules is likely to yield gradual improvement or encounter abrupt changes.

Both panels were produced with the same enumeration and projection procedure, and the vertical scale represents energy in the same units, enabling qualitative comparison of landscape shape and the relative prevalence of low‑energy areas across the two task graphs.}
    \label{fig:search_space_landscape}
\end{figure}

Figure~\ref{fig:search_space_landscape} shows how workflow structure changes the scheduling problem. wide DAGs create a rugged landscape with shallow valleys everywhere. When many tasks are ready at once, there are countless ways to assign them to VMs. A small change in the assignment can cause energy consumption to jump around unpredictably.

LongCP DAGs produce a different landscape entirely. Fewer valleys, but deeper ones. The critical path constrains most decisions because dependency chains force a specific order. There's little room to explore alternatives. The main freedom is placing short side branches, which creates a few isolated basins instead of a chaotic surface.

With these distinctions, we address \hyperref[q:joint-influence]{Q1} and \hyperref[q:cross-generalization]{Q2} through four host-configuration regimes (AL, NA, HS, HP) in our queue-free setting. 


\subsubsection*{Queue-Free Regime}
\label{sec:queue_free_shared}

Across all four regimes we assume a \emph{queue-free}, single-workflow
environment. A cluster of $V$ virtual machines (VMs) executes a single DAG
workflow. Time is continuous and tasks are non-migratable once assigned to a
VM.

The dataset generator enforces queue-freedom: for each DAG we scale task memory
and CPU requirements so that the peak-width layer fits within aggregate cluster
capacity. If $\mathcal{L}_{\max}$ is the peak layer,
\[
  \sum_{i \in \mathcal{L}_{\max}} \text{req\_mem}_i 
  \;\le\; \sum_{v=1}^V \text{mem}_v,
  \quad
  \sum_{i \in \mathcal{L}_{\max}} \text{req\_cores}_i 
  \;\le\; \sum_{v=1}^V \text{cores}_v.
\]
Thus, whenever a task becomes ready, there exists a feasible placement that
does not violate capacity. The four regimes below differ only in how
VM \emph{speed} and \emph{power} are parameterized.

\subsubsection*{Homogeneous-Speed Regime (HS)}
\label{sec:queue_free_homogeneous_speed}

In the homogeneous-speed regime, all VMs process work at the same rate $s$ (e.g., MIPS). If a task $i$ has computational length $L_i$ (in MI), its processing time is $\tau_i = L_i / s$, independent of which VM executes it.

Because speeds are identical and the instance is queue-free, any non-pathological schedule that keeps critical-path tasks busy attains the same makespan $T^\star = L_{\mathrm{CP}} / s$, where $L_{\mathrm{CP}}$ is the critical-path length of the DAG. Makespan is determined by DAG topology, not by VM assignment choices.

We focus on power-heterogeneous but speed-homogeneous hosts. At approximately fixed makespan, the only remaining degree of freedom is which VMs are active and for how long. This means the policy can only shape active energy by controlling VM power profiles.


\subsubsection*{Homogeneous-Power Regime (HP)}
\label{sec:queue_free_homogeneous_power}

In the homogeneous-power regime, all VMs share the same active power $P^{\mathrm{act}}$ but differ in speed $s(v)$. Active energy is approximated as $E \approx \int P^{\mathrm{act}}\,dt$. For a fixed task workload, this makes energy largely insensitive to speed (ignoring second-order utilization effects), while makespan still depends on $s(v)$ through $\tau = L/s(v)$.

The effective objective becomes primarily time-dominated: faster machines shorten makespan, but active energy remains approximately unchanged across VM choices since $P^{\mathrm{act}}$ is constant. This reduces the multi-objective problem to a single-objective problem focused on minimizing completion time.

\subsubsection*{Heterogeneous Aligned Regime (AL)}
\label{sec:queue_free_heterogeneous_aligned}

In the heterogeneous aligned regime, speed and energy efficiency move in the same direction. Faster machines are also more power efficient for the same amount of work. Formally, for two VMs $v_1$ and $v_2$ with speeds $s(v_1) < s(v_2)$, the faster VM has lower energy. Moving a task to a faster VM therefore tends to reduce both completion time and total energy, so the two objectives are mostly aligned.


\subsubsection*{Heterogeneous Non Aligned Regime (NA)}
\label{sec:queue_free_heterogeneous_nonaligned}

In the heterogeneous non aligned regime, we break this link between speed and efficiency. Some VMs are fast but energy hungry, while others are slow but energy cheap. In other words, higher speed does not imply lower energy per unit work, and in some cases it can be strictly worse.

This creates real conflicts between local choices. A faster VM may reduce completion time but increase total energy, while a slower and more efficient VM may save energy at the cost of longer makespan. The trade off surface becomes non monotone, with several locally attractive choices depending on the exact speed and power pairing.



\begin{figure}[htp]
    \centering
    \includegraphics[width=0.7\linewidth]{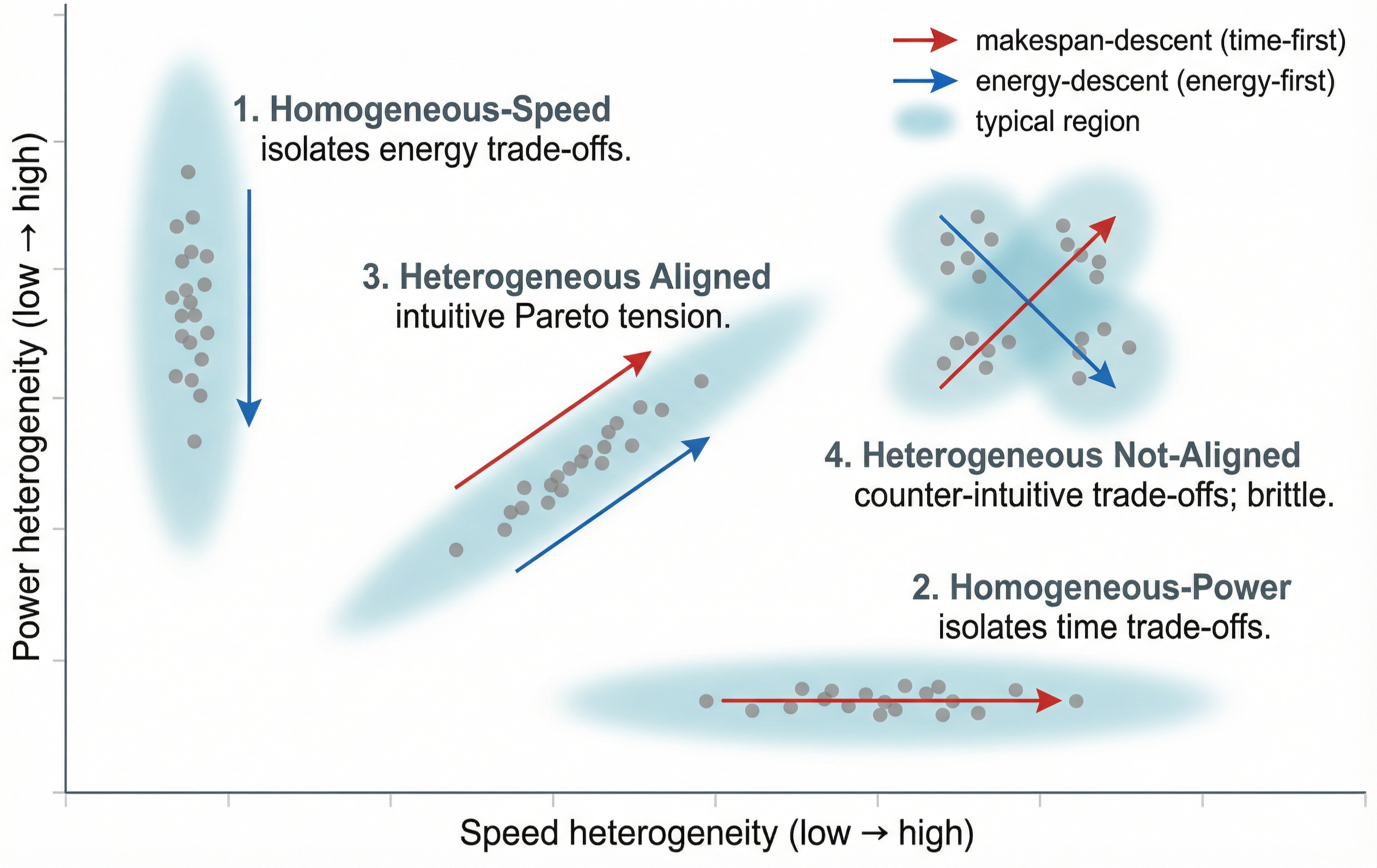}
    \caption{\textbf{Speed--power host regimes.}
    Illustration of the four queue-free host regimes studied in this paper.
    Homogeneous-Speed: all VMs share the same processing speed but differ in
    power, isolating energy trade-offs at fixed makespan.
    Homogeneous-Power: all VMs share the same active power but differ in
    speed, isolating time trade-offs.
    Heterogeneous Aligned: speed and power are monotonically aligned, inducing
    an intuitive Pareto frontier between makespan and active energy.
    Heterogeneous Non-Aligned: speed and power are not aligned, creating
    counter-intuitive trade-offs and stronger generalization challenges for
    learned schedulers.}
    \Description{A four quadrant schematic compares queue free host regimes by how virtual machines differ in processing speed and active power. Each quadrant shows a small set of VM markers that illustrate the relation between speed and power.

Homogeneous Speed: all VMs have the same speed and different power. For a fixed workload the makespan is the same regardless of which VM is chosen, while total active energy changes with the selected power levels.

Homogeneous Power: all VMs have the same active power and different speeds. Power per unit time is constant. Faster VMs shorten runtime and reduce active energy, slower VMs lengthen runtime and increase energy.

Heterogeneous Aligned: speed and power increase together. Faster VMs always draw more power. This yields a clear trade off between time and energy, forming an intuitive Pareto frontier from slow low power to fast high power choices.

Heterogeneous Non Aligned: speed and power have no consistent order. Some VMs are fast and low power, others are slow and high power. Dominance is unclear and the Pareto set can be irregular, creating counterintuitive trade offs and tougher generalization for learned schedulers.
}
    \label{fig:host_regime_compass}
\end{figure}

%

\section{Model Architecture}

\label{sec:architecture}

\begin{figure*}[htp]
    \centering
    \includegraphics[width=0.8\linewidth]{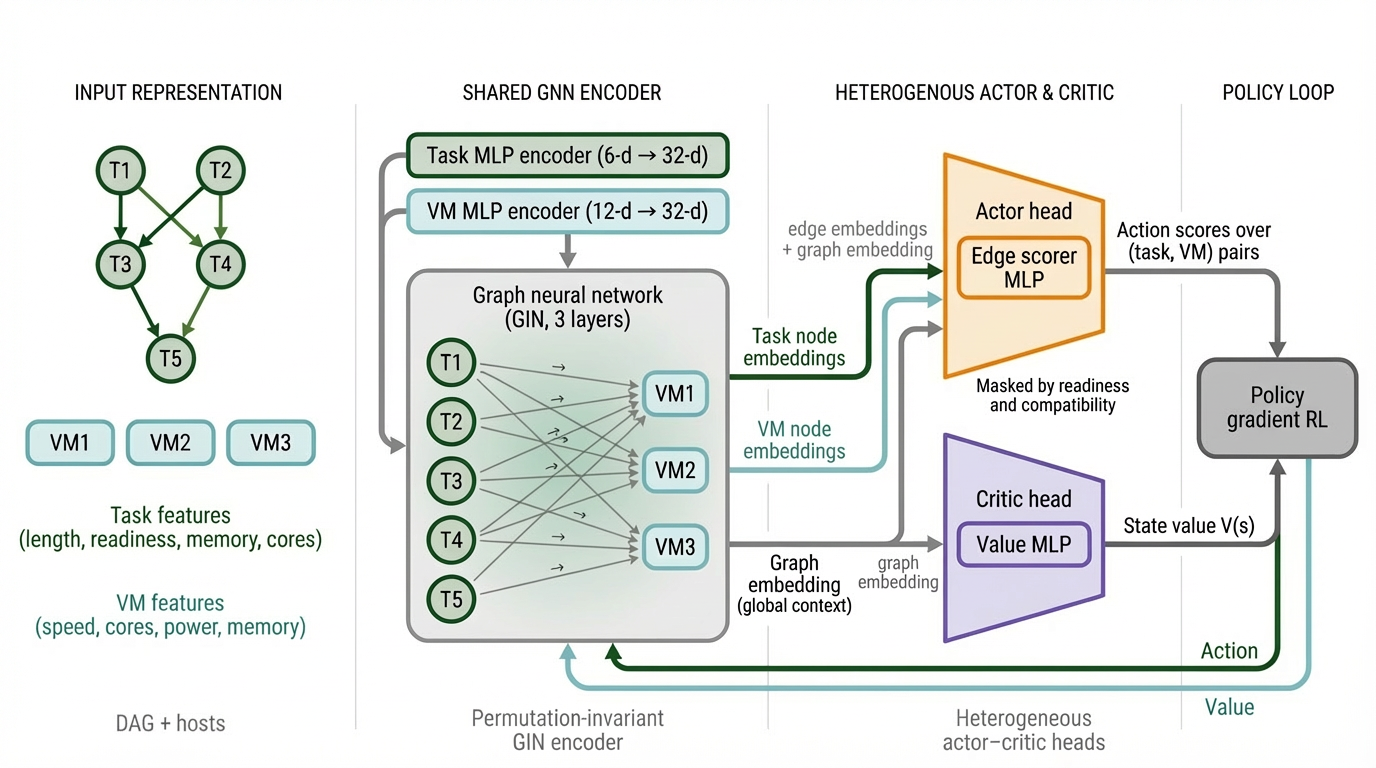}
    \caption{Overview of the proposed GIN-based actor--critic scheduler architecture.}
    \Description{Block diagram of a scheduler that uses an actor and a critic with a shared graph neural network. Inputs are two sets of nodes and two edge types. Task nodes carry status flags, remaining work, completion time, and resource needs. VM nodes carry current load, completion times, speed, core counts, available cores, memory capacity, free memory, and idle and peak power. Dependency edges connect parent and child tasks. Compatibility edges connect tasks to VMs that can run them.

Two small multilayer perceptrons encode tasks and VMs into a common latent space. Encoded nodes pass through several Graph Isomorphism Network layers that aggregate over both edge families. A global pooling produces a single graph embedding shared by both heads.

The actor head builds an embedding for each feasible task and VM pair by combining the task node, the VM node, and the graph embedding. A small scorer maps each pair to a scalar. Invalid pairs are masked. The remaining scores form a task by VM matrix, flattened and passed through softmax to produce a probability distribution over actions.

The critic head takes only the graph embedding and uses a small multilayer perceptron to output a single value that estimates expected return from the current state. Arrows indicate shared parameters between encoders and backbone, and separate outputs for policy and value.}
    \label{fig:gnn_architecture}
\end{figure*}

Our scheduler is a deep actor--critic architecture built around a shared
graph neural network (GNN) backbone that embeds both workflow tasks and virtual
machines (VMs). Figure~\ref{fig:gnn_architecture} summarizes the deep reinforcement learning
architecture used throughout this work. We follow the standard actor--critic
decomposition : an actor (policy network) outputs
a stochastic policy $\pi_\theta(a \mid s)$ over scheduling actions, while a critic
(value network) estimates the state value $V_\phi(s)$ and is used as a learned
baseline to reduce the variance of the policy-gradient estimator
(\cite{konda1999actor,mnih2016asynchronous,schulman2017proximal}). Both heads
share the same Graph Isomorphism Network (GIN) backbone so that policy and value are learned from a common
structural representation of the workflow and VM pool.

\paragraph{Input representation.}
At each decision step the environment provides a structured observation
encoding the current workflow DAG and the VM pool. Tasks are represented by:
(i) scheduled/ready flags, (ii) remaining work and completion time,
(iii) CPU and memory requirements. VMs are represented by: (i) completion
times and current utilization, (ii) speed, core count and available cores,
(iii) memory capacity and free memory, (iv) host idle and peak power.
Compatibility edges link tasks to VMs on which they can run, and dependency
edges encode the workflow DAG (parent--child relations). The resulting structure forms a bipartite-plus-dependency graph with heterogeneous node types and two distinct edge families.

\paragraph{Task and VM encoders.}
We map raw task and VM features into a common latent space using two separate
multi-layer perceptrons (MLPs). The task encoder consumes a 6-dimensional
feature vector and outputs a $d$-dimensional embedding. The VM encoder
consumes a 12-dimensional feature vector and outputs an embedding in the same
space. Both encoders use batch normalization and ReLU nonlinearities, followed
by a final linear projection. The encoders are shared between actor and
critic.

\paragraph{GIN backbone.}
We concatenate the encoded task and VM nodes into a single set and process them using a three-layer Graph Isomorphism Network (GIN) (\cite{xu2019how}) with hidden dimension $h$. The edges capture both task–VM compatibility links and task–task dependency links. Each GIN layer performs neighborhood aggregation followed by an MLP update, producing type-agnostic node embeddings that capture both the workflow structure and the VM context. A global mean-pooling operation over all nodes is then applied to compute a \emph{graph embedding}, providing a compact summary of the current scheduling state.

\paragraph{Actor head.}
The actor network uses the learned node and graph embeddings to evaluate possible scheduling decisions. 
For each task that can run on a given VM, we build an \emph{edge embedding} by combining the embeddings of the task node, the VM node, and the overall graph representation. 
This combined vector is passed through a small multilayer perceptron (the \emph{edge scorer}) that outputs a single numerical score. 
All scores are then arranged into a task$\times$VM matrix, where invalid entries that corresponds to tasks that are not ready, already scheduled, or incompatible are masked out. 
After flattening the remaining entries, a softmax function converts the scores into a probability distribution over all valid $(\text{task}, \text{VM})$ pairs. 
The agent’s policy $\pi_\theta(a \mid s)$ is defined by sampling or selecting the highest-scoring action from this distribution.

\paragraph{Critic head.}
The critic shares the same GIN backbone and encoders but operates only on the
global graph embedding. A two-layer MLP maps the graph embedding to a scalar
value estimate $V_\phi(s)$, representing the expected return from the current
state under the policy. 

\paragraph{Training.}
Actor and critic parameters are jointly optimized with a policy-gradient
algorithm using the mixed objective described in
Section~\ref{par:reward}. The actor receives policy-gradient updates
based on advantage estimates that combine makespan and active-energy,
while the critic is trained with a regression loss toward bootstrapped returns. Gradients flow through the encoders and GIN backbone,
so the resulting node, edge, and graph embeddings become specialized for
topology-aware scheduling decisions across the host regimes described in the next section.

\section{Experimental Methodology, Results and Analysis}

\label{sec:exp_methodology}

This section shows the experiments and results that really answer Q1 (how DAG topology and host setups affect scheduling choices) and Q2 (how well policies work across different topologies).

\subsection{Experimental Methodology}

We conduct experiments using the two workflow families (wide and LongCP) and four host regimes (HS, HP, AL, NA) defined in Section~\ref{sec:mostimportant}. For each topology class, we define disjoint training and evaluation seed sets to generate independent workflow instances.

\subsubsection{Training Setup}
\label{sec:exp_training}

We train the GNN-based actor--critic agent described in Section~\ref{sec:architecture} using standard on-policy policy gradients with advantage estimation.

\paragraph{Episodes and environment dynamics.}
Each episode executes a complete workflow from source tasks to sink completion. The environment implements the transition kernel $\mathcal{P}$ from Section~\ref{sec:PF}, simulating task execution, updating readiness based on DAG dependencies, and integrating power over time.

\paragraph{Reward formulation.}
We use the mixed objective from Section~\ref{par:reward}:
\[
r_k = w_T \cdot \Delta R^{\text{mk}}_k + w_E \cdot \Delta R^{\text{en}}_k
\]
with per-decision incremental improvements in makespan and energy. Coefficients $(w_T, w_E)$ are fixed as (1,1)  within experiments.

\paragraph{Training distributions.}
For each host regime, we train two specialist policies:
\begin{enumerate}
  \item \textbf{wide-only:} episodes contain only 10 wide workflows.
  \item \textbf{LongCP-only:} episodes contain only 10 LongCP workflows.

\end{enumerate}

\paragraph{Training hyperparameters.}
All agents use the same PPO configuration: $2{,}000{,}000$ total timesteps with $10$ parallel environments, batch size $2560$ ($256$ steps per environment), $4$ minibatches, $4$ update epochs per batch, learning rate $2.5\times 10^{-4}$, GAE with $\gamma=0.99$ and $\lambda=0.95$, and clip coefficient $0.2$. 


\subsubsection{Evaluation Protocol}
\label{sec:exp_evaluation}

\paragraph{In-distribution and cross-topology evaluations.}
For each trained agent, we evaluate under all combinations of:
\begin{itemize}
  \item training topology: wide-only vs.\ LongCP-only.,
  \item test topology: wide vs.\ LongCP,
  \item host regime: HS, HP, AL, NA.
\end{itemize}
This produces in-distribution conditions (e.g., wide-trained agent on
wide workflows) as well as cross-topology conditions (e.g.,
wide-trained agent on LongCP workflows), under each host regime.

\paragraph{Metrics.}
For every configuration we report:
\begin{itemize}
  \item average makespan per workflow.
  \item average active energy per workflow;
  \item the empirical Pareto relationship between energy and makespan
        across different agents and baselines.
\end{itemize}


\subsection{Results and Analysis}
\label{Sec:Results}
\subsubsection{Results}

Table~\ref{tab:hetero-all-hosts} summarizes cross-domain performance of the wide and LongCP heterogeneous specialists across the HS, HP, AL, and NA host configurations.
\subsection*{Homogeneous-Speed (HS)}
All VMs have the same speed, so makespan is driven by DAG topology and keeping the critical path busy. Both specialists match makespan on their evaluation domains (2.08 on LongCP, 0.70 on wide). The wide specialist uses less active energy in both domains (51.00 vs 52.87 on LongCP, 57.11 vs 58.92 on wide). This means the wide specialist wins on energy while maintaining the same makespan, even when tested on LongCP configurations where it was never trained. The reason is that training on wide parallel structures teaches the agent to spread work efficiently across VMs with different power profiles. 

\subsection*{Homogeneous-Power (HP)}
All VMs share the same power, but speeds vary from 160 to 800~GIPS. Energy becomes time-driven. The LongCP specialist is faster on both domains (3.48 vs 3.62 on LongCP, 1.26 vs 1.27 on wide). Energy is essentially tied on LongCP (21.76 vs 21.76) and on wide (24.33 vs 24.34). This happens because training on long dependency chains teaches the agent to prioritize critical path optimization and assign important tasks to faster VMs. When power is fixed, this speed-aware scheduling becomes the dominant factor for both makespan and energy. The wide specialist, trained on more parallel structures, learns a more exploratory policy that spreads work broadly but misses the structured prioritization that matters when speed varies. 

\subsection*{Heterogeneous Aligned (AL)}
Speed and power are aligned: faster VMs are more energy-efficient. Using the same speed distribution as HP and NA, the wide specialist dominates on both time and energy in both domains: LongCP eval 4.05 vs 4.07 makespan and 50.91 vs 51.58 energy, wide eval 1.37 vs 1.39 makespan and 56.98 vs 58.31 energy. When speed aligns with efficiency, the wide policy’s tendency to keep more work on faster machines helps both makespan and energy.

\subsection*{Heterogeneous Non-Aligned (NA)}
Fast VMs are power-hungry and slow VMs are efficient. Using the same speed distribution, a real tradeoff appears. On LongCP eval, LongCP is faster (3.04 vs 3.14) while wide uses less energy (50.87 vs 52.34). On wide eval, LongCP is slightly faster (1.17 vs 1.18) and higher energy (58.45 vs 56.72). 

 \begin{table*}[htp]
\caption{Cross-domain evaluation of heterogeneous agents across host configurations.
         Best results for each evaluation domain within a host configuration are in bold.}
\label{tab:hetero-all-hosts}
\centering
\small
\begin{tabular}{@{} lll rr @{}}
\toprule
Host cfg & Method & Eval Domain & Makespan & \makecell{Active\\Energy} \\
\midrule

\multicolumn{5}{l}{\itshape HS host configuration} \\
\rowcolor{blue!8}
HS & LongCP & LongCP & \textbf{2.08} & 52.87 \\
\rowcolor{green!8}
HS & wide   & LongCP & \textbf{2.08} & \textbf{51.00} \\

\rowcolor{blue!8}
HS & LongCP & wide   & \textbf{0.70} & 58.92 \\
\rowcolor{green!8}
HS & wide   & wide   & \textbf{0.70} & \textbf{57.11} \\
\addlinespace[0.6em]

\multicolumn{5}{l}{\itshape HP host configuration} \\
\rowcolor{blue!8}
HP & LongCP & LongCP & \textbf{3.48} & \textbf{21.76} \\
\rowcolor{green!8}
HP & wide   & LongCP & 3.62 & \textbf{21.76} \\

\rowcolor{blue!8}
HP & LongCP & wide   & \textbf{1.26} & \textbf{24.34} \\
\rowcolor{green!8}
HP & wide   & wide   & 1.27 & \textbf{24.34} \\
\addlinespace[0.6em]

\multicolumn{5}{l}{\itshape AL host configuration} \\
\rowcolor{blue!8}
AL & LongCP & LongCP & 4.07 & 51.58 \\
\rowcolor{green!8}
AL & wide   & LongCP & \textbf{4.05} & \textbf{50.91} \\

\rowcolor{blue!8}
AL & LongCP & wide   & 1.39 & 58.31 \\
\rowcolor{green!8}
AL & wide   & wide   & \textbf{1.37} & \textbf{56.98} \\
\addlinespace[0.6em]

\multicolumn{5}{l}{\itshape NA host configuration} \\
\rowcolor{blue!8}
NA & LongCP & LongCP & \textbf{3.04} & 52.34 \\
\rowcolor{green!8}
NA & wide   & LongCP & 3.14 & \textbf{50.87} \\

\rowcolor{blue!8}
NA & LongCP & wide   & \textbf{1.17} & 58.45 \\
\rowcolor{green!8}
NA & wide   & wide   & 1.18 & \textbf{56.72} \\

\bottomrule
\end{tabular}
\end{table*}



\begin{figure}[htpb]
  \centering
  \includegraphics[width=0.7\textwidth]{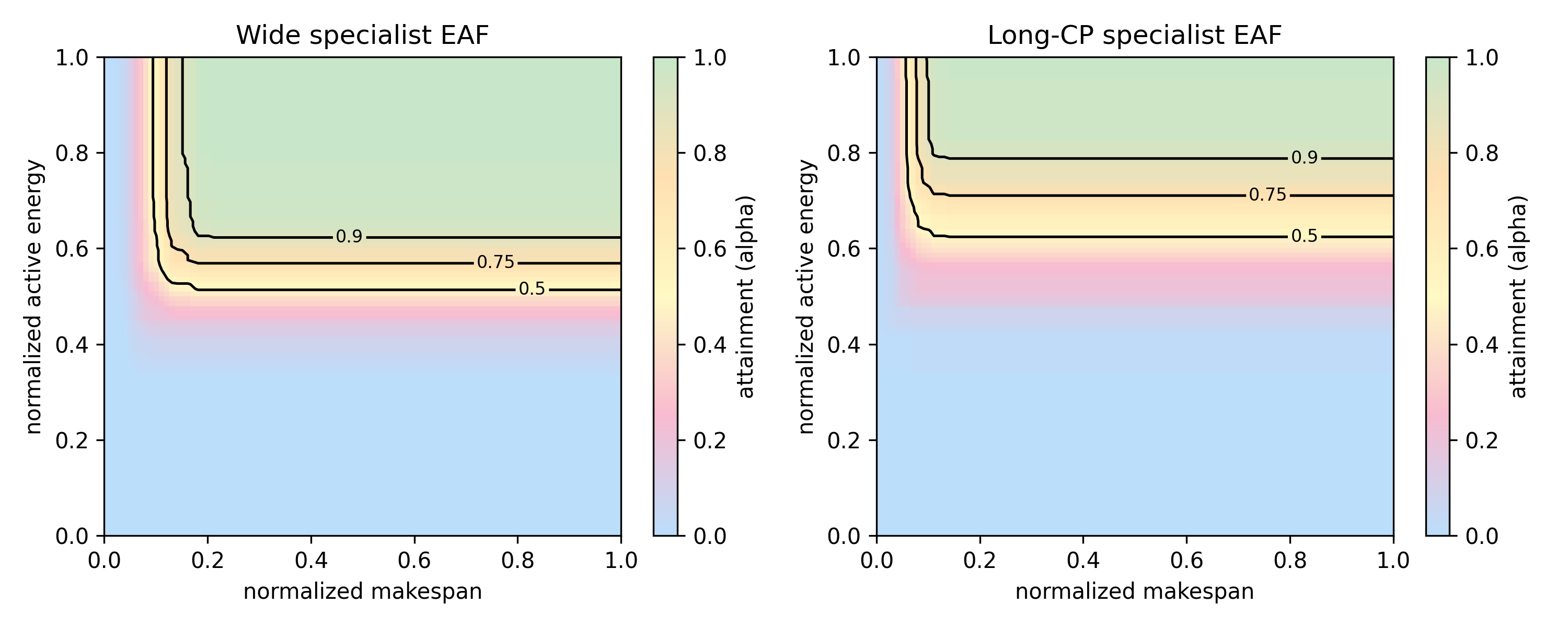}\\[0.8em]
  \includegraphics[width=0.7\textwidth]{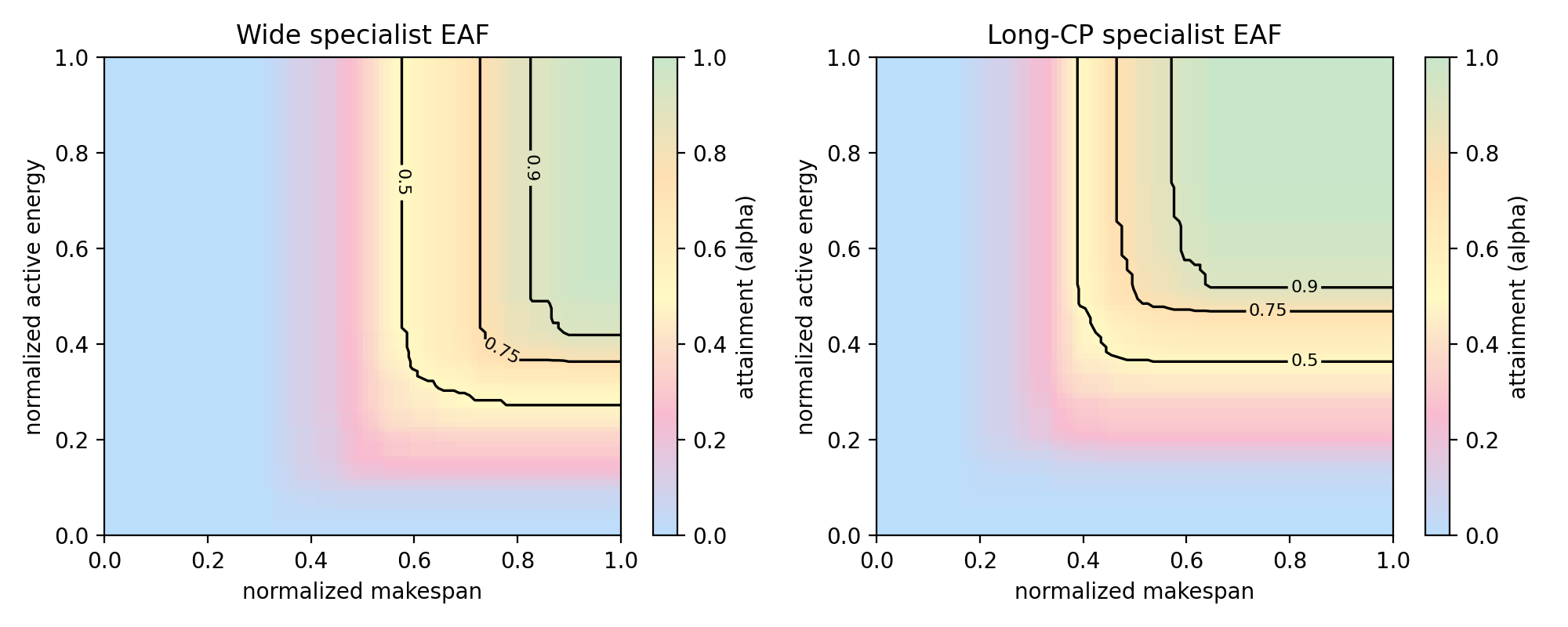}
  \caption{
  Empirical attainment functions (EAFs) over 100 test jobs for the wide and LongCP specialists on the wide configuration (top) and the LongCP configuration (bottom). Each panel shows the distribution of the trade-off between normalized makespan (x-axis) and normalized active energy (y-axis), with color indicating the attainment level~$\alpha$ and black contours marking $\alpha \in \{0.5, 0.75, 0.9\}$.
  }
  \Description{Two stacked panels summarize empirical attainment over 100 test jobs for two schedulers: a LongCP specialist and a wide specialist. Axes show normalized makespan on the horizontal axis and normalized active energy on the vertical axis. Each panel depicts the empirical attainment function as filled levels with overlaid contours at attainment levels 0.5, 0.75, and 0.9. The top panel uses the wide host configuration; the bottom panel uses the LongCP configuration. In each panel, the shapes of the attainment regions indicate how often a scheduler reaches points in the time–energy plane. Regions shifted left indicate shorter makespan, regions shifted down indicate lower energy. The side by side attainment patterns allow comparison of which scheduler more frequently reaches low time versus low energy regions, and how the full distribution of outcomes spreads across the trade off frontier.}

  \label{fig:eaf_specialists}
\end{figure}

\autoref{fig:eaf_specialists} confirms the headline trade-offs: the LongCP specialist tends to dominate the low-makespan region, while the wide specialist tends to dominate the low-energy region. This mirrors the averages and shows the biases across the whole distribution.
\begin{figure}[htpb]
  \centering
  \includegraphics[width=0.7\linewidth]{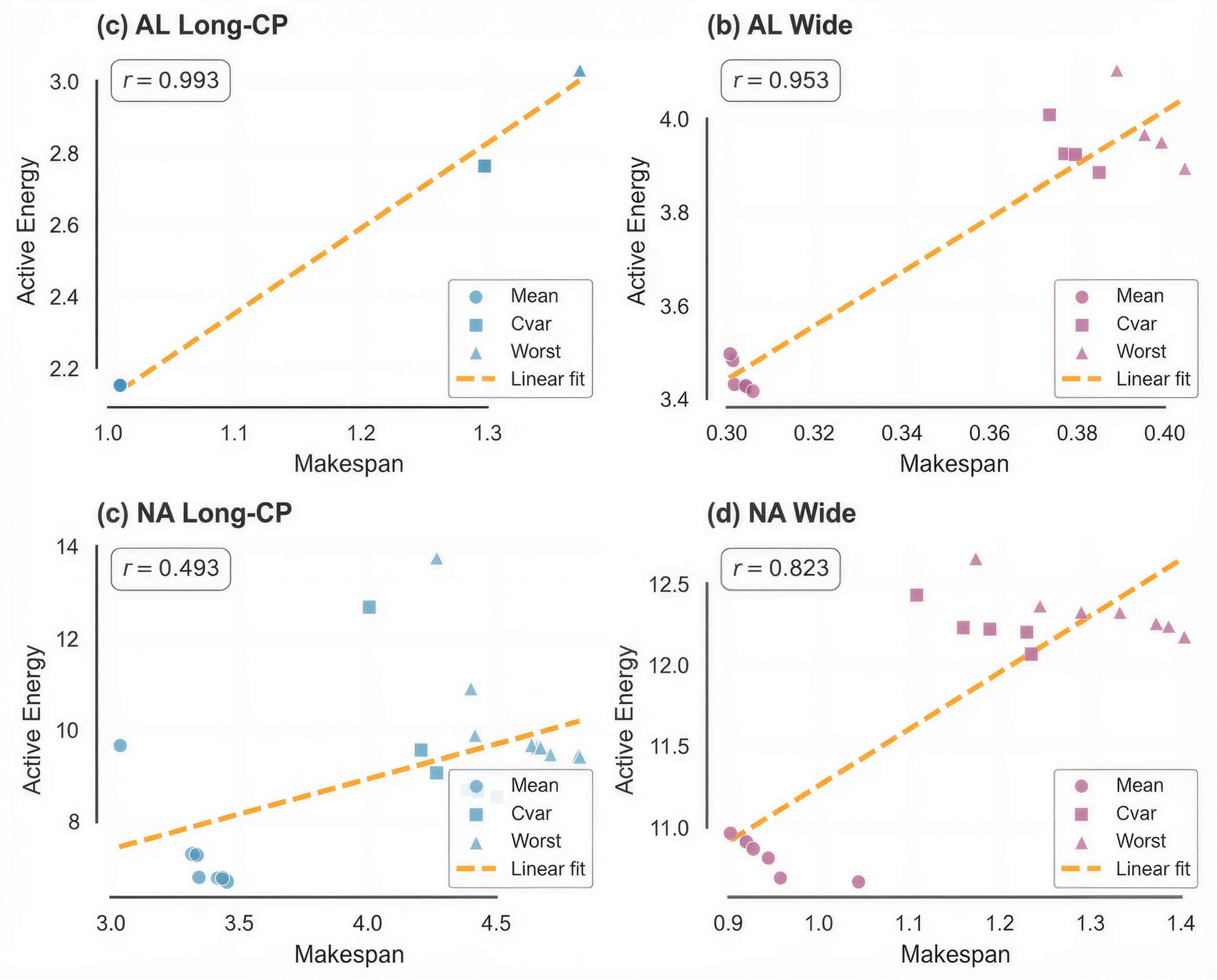}
  \caption{Correlation between makespan and active energy across Pareto checkpoints
  for AL (top row) and NA (bottom row), with LongCP and wide specialists.}
  \Description{Grid of subplots reporting the correlation between makespan and active energy across saved Pareto checkpoints. The top row corresponds to an aligned regime and the bottom row to a non aligned regime. Columns compare LongCP and wide specialists. Each subplot summarizes, for a sequence of model checkpoints along the Pareto front, how strongly time and energy move together. Positive correlation means runs with longer time also tend to use more energy; negative correlation means improvements in time often come with higher energy or vice versa, indicating a stronger trade off. Near zero indicates weak or inconsistent relationship. The layout lets the reader see how regime and specialization affect the sign and magnitude of the objective correlation across training checkpoints.}

  \label{fig:objective_correlation_na_al}
\end{figure}

The diagnostic referenced earlier now appears in \autoref{fig:objective_correlation_na_al}. Panel (a) (AL, Long‑CP) shows near‑perfect coupling ($r{=}0.993$): makespan and active energy move almost linearly together. Panel (b) (AL, wide) remains strongly coupled ($r{=}0.953$), though with a slightly wider band. Now , under NA, the link weakens: panel (c) (NA, Long‑CP) shows only moderate correlation with substantial spread ($r{=}0.493$), while panel (d) (NA, wide) is stronger but still looser than AL ($r{=}0.823$) and exhibits more variance at lower makespans. 

AL creates a strong link between time and energy, so improving one usually improves the other. NA loosens this connection, especially for Long CP workflows, allowing real trade offs between time and energy. This pattern also shows up in the EAFs: under NA, different specialists dominate different parts of the frontier, rather than a single agent leading everywhere

\begin{figure}[htp]
  \centering
  \includegraphics[width=0.7\linewidth]{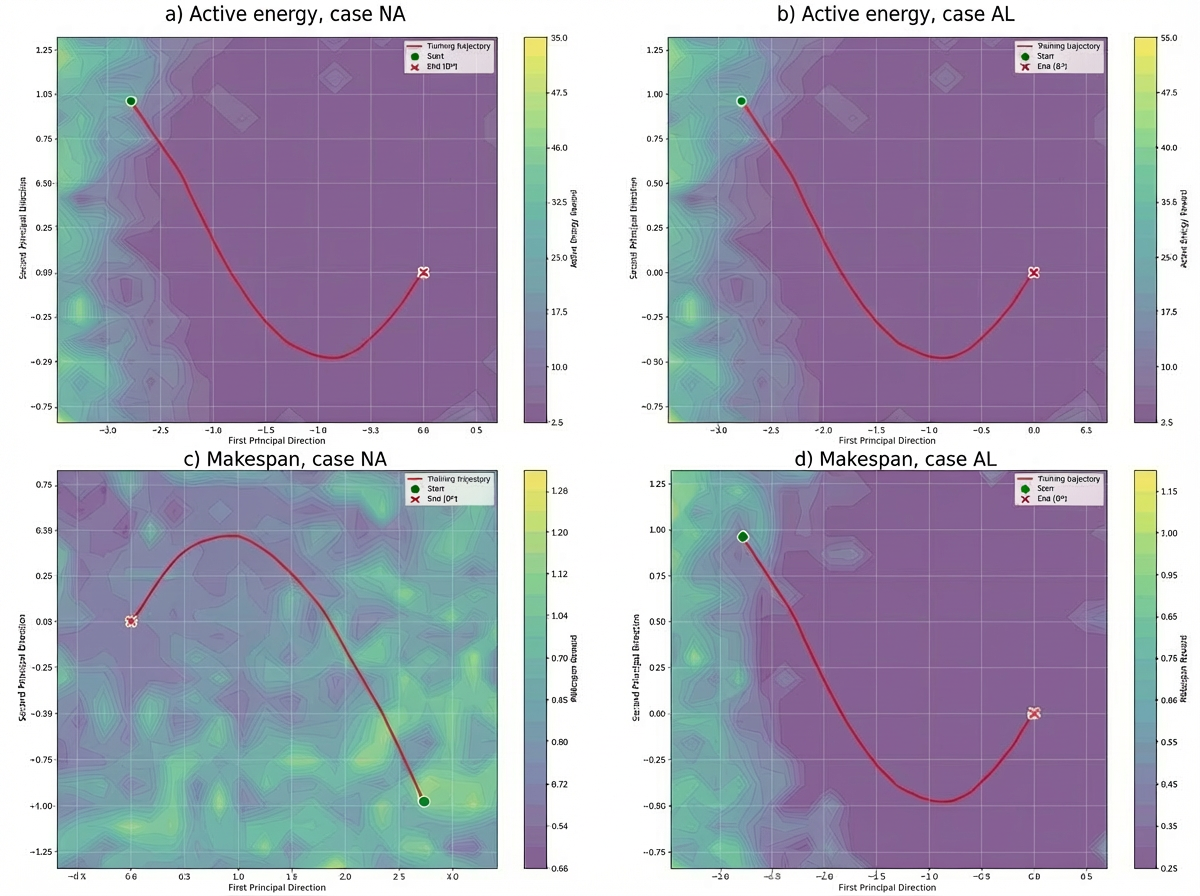}

  \caption{Panels: a) Active energy, case NA; b) Active energy, case AL; c) Makespan, case NA; d) Makespan, case AL.}
  \Description{A 2 by 2 layout shows objective landscapes in a 2D projection of the actor’s parameter space, with the same axes across panels labeled First Principal Direction and Second Principal Direction. Each panel displays a background field representing the objective value over this plane and a single training trajectory drawn as a curve with a start marker (filled circle) and an end marker (X). Lower values are better for both objectives.

Panel a (active energy, non aligned regime): the trajectory starts near the upper left and moves downhill toward a broad low energy basin near the center left, then continues to the right, ending slightly above the basin’s minimum.

Panel b (active energy, aligned regime): the path again descends toward a central low energy region but finishes to the right of the minimum, indicating a modest rise relative to the best area.

Panel c (makespan, non aligned regime): the trajectory begins left of center, ascends mildly, then turns and descends toward the lower right, finishing close to a region with low makespan.

Panel d (makespan, aligned regime): the path starts near the upper left, curves toward a low basin near the lower center, and ends mid right, not at the lowest area.

Across panels, the same projected coordinates allow comparison of how training moves through different landscapes under two regimes (aligned and non aligned) and two objectives (active energy and makespan). A colorbar is present to indicate objective magnitude, while the curve and markers highlight the progression from initialization to the final checkpoint.}
    \label{fig:actor_landscape}
\end{figure}

These patterns raise a natural question: why do two agents trained with the same mixed objective end up with such different preferences over time and energy? The answer lies in how DAG topology shapes which states the agent visits during training.

Let $\pi$ be a policy over observations $s$ (task requirements and readiness, VM utilization, energy rate features), and let $d_\pi(s)$ denote its state occupancy measure. Both agents are trained with the same mixed objective described in Section~\ref{sec:PF}:
\[
J(\pi) = \alpha\,\mathbb{E}[E_{\text{active}}] + \beta\,\mathbb{E}[\text{Makespan}], \qquad \alpha,\beta>0,
\]
but they visit different parts of the state space because of their training DAGs.

\begin{figure}[htp]
  \centering
  \includegraphics[width=0.7\linewidth]{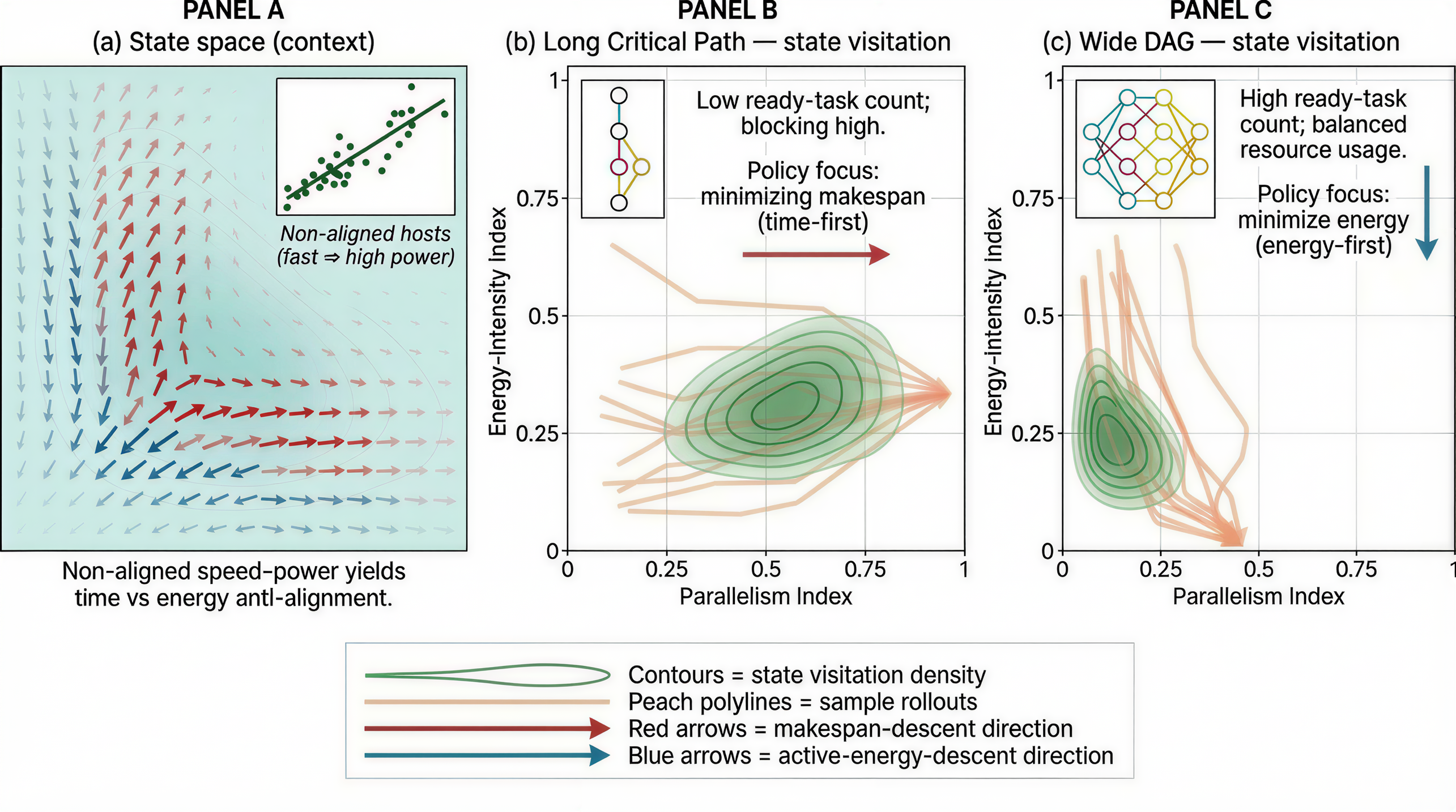}
  \caption{\textbf{State visitation under a fixed mixed objective and non-aligned speed–power.}
  (a) Conceptual state-space diagram with anti-aligned red (makespan) and blue (energy) descent directions, plus an inset showing non-aligned host speed–power. (b) Long Critical Path (LongCP) training: KDE contours and rollouts concentrate in high-parallelism regions. (c) wide DAG training: visitation shifts toward lower-parallelism states.}
  \label{fig:state-visitation-structure}
  \Description{A 2 by 2 layout shows objective landscapes in a 2D projection of the actor’s parameter space, with the same axes across panels labeled First Principal Direction and Second Principal Direction. Each panel displays a background field representing the objective value over this plane and a single training trajectory drawn as a curve with a start marker (filled circle) and an end marker (X). Lower values are better for both objectives.

Panel a (active energy, non aligned regime): the trajectory starts near the upper left and moves downhill toward a broad low energy basin near the center left, then continues to the right, ending slightly above the basin’s minimum.

Panel b (active energy, aligned regime): the path again descends toward a central low energy region but finishes to the right of the minimum, indicating a modest rise relative to the best area.

Panel c (makespan, non aligned regime): the trajectory begins left of center, ascends mildly, then turns and descends toward the lower right, finishing close to a region with low makespan.

Panel d (makespan, aligned regime): the path starts near the upper left, curves toward a low basin near the lower center, and ends mid right, not at the lowest area.

Across panels, the same projected coordinates allow comparison of how training moves through different landscapes under two regimes (aligned and non aligned) and two objectives (active energy and makespan). A colorbar is present to indicate objective magnitude, while the curve and markers highlight the progression from initialization to the final checkpoint.}
\end{figure}

Figure~\ref{fig:state-visitation-structure} shows how this works. The x-axis shows a parallelism index (normalized ready tasks per level / DAG width) and the y-axis shows an energy intensity index (aggregate active power rate while busy).

Crucially, both agents optimize the same $J(\pi)$. The difference is that DAG topology changes which states are reachable and frequently visited. Long critical paths (Panel~\ref{fig:state-visitation-structure}b) compress the feasible manifold toward low parallelism, so $d_\pi$ places more mass in low-parallelism regions where energy-reducing choices matter. wide DAGs (Panel~\ref{fig:state-visitation-structure}c) expand the feasible trajectory set $\mathcal{T}(\text{DAG},\text{hosts})$ in parallel directions, so $d_\pi$ shifts toward high-parallelism states where time-reducing choices are available. Structure rotates which parts of the trade-off surface are accessible during training.

Figure~\ref{fig:actor_landscape} provides additional evidence by showing how the learned value landscapes change with actor parameters during training. In the aligned case (AL), both makespan and energy gradients point in similar directions because faster VMs also consume less power. In the non-aligned case (NA), the optimal regions for makespan and active energy need not coincide: improving one does not necessarily improve the other, and the two objectives can favor different parts of parameter space. This reflects the different biases learned from different $d_\pi(s)$ patterns.

This explains the cross-evaluation results in Table~\ref{tab:hetero-all-hosts}. When the LongCP specialist is tested on wide configurations, it still tries to minimize makespan. When the wide specialist is tested on LongCP configurations, it still tries to minimize energy because that's what it learned to prioritize when it had choices. Neither strategy is wrong, they just learned different priorities from different training distributions. The DAG topology during training fundamentally shapes which features the policy learns to care about, even when the objective function stays the same. Policy gradients are weighted by $d_\pi(s)$, so topology and host regime determine which features and transitions are emphasized during learning.

This is the regime where domain-specific training matters most: choosing the right specialist depends on whether the deployment prioritizes time or energy.

\section{Conclusion}

The challenge of efficiently scheduling complex, dependency-driven workflows in modern cloud environments requires balancing completion time (makespan) against energy consumption. Deep Reinforcement Learning (deep RL) schedulers with Graph Neural Network (GNN) backbones are a promising alternative to classical heuristics, but their robustness and generalization across workflow structures and hardware regimes remain open questions. In this work, we addressed this by constructing a controlled benchmark that factorizes the problem into two contrasting DAG topologies : wide Parallel and Long Critical Path (LongCP) , and four host regimes: Homogeneous Speed (HS), Homogeneous Power (HP), Aligned (AL), and Non-Aligned (NA). By training specialized agents and subjecting them to systematic cross-topology and cross-regime evaluation, we have successfully answered the two research questions posed in Section~\ref{sec:introduction} and made several key contributions to the understanding of learned cloud scheduling: 

Our study shows that DAG topology and host regime shape what deep RL schedulers actually learn to prioritize, even under the same energy–makespan objective. With homogeneous speed (HS), makespan is the same across specialists on each domain, and the wide specialist consistently uses less active energy while matching time. With homogeneous power (HP), the LongCP specialist is faster on both domains. In the aligned regime (AL), where faster machines are also more efficient, the wide specialist wins on both makespan and energy across both domains. In the non-aligned regime (NA), a real tradeoff appears: LongCP achieves better makespan while wide achieves lower energy. In brief, there is no single best specialist. Each policy acts like a regime‑specific heuristic: LongCP pushes time down via critical‑path focus, wide reduces energy by distributing work, and the preferred choice depends on the deployment’s speed–power correlation.

This structural and regime dependence matters for production. A single agent will not cover all workflows and hardware. Different combinations call for different strategies. LongCP works well when power is uniform or aligned with speed because it focuses on the critical path. wide saves energy when power favors spreading work because it uses parallelism and load distribution. Knowing which structures and power relations each specialist exploits helps predict performance and pick the right policy. It also guides data curation for a larger agent by telling us which jobs to include so the training mix matches the target regimes.

This opens clear next steps. We should identify which graph features and hardware traits the agents attend to using attribution on actions or embeddings. This will improve interpretability and help predict transfer. We should also test domain agnostic training such as meta learning or domain randomization that pushes the agent to learn features that hold across topologies and host regimes, aiming for a single robust policy that adapts at run time.

Finally, our study only looked at a single workflow with no queue. Real systems are more complex. They run many workflows at once and must deal with contention and priorities. Extending this analysis to a queue based setting is essential. In that case, DAG topology, system load, and hardware heterogeneity will interact in new
ways. A critical-path policy may behave differently when multiple long chains compete for the same fast machines.
A parallelism-aware policy may struggle when the queue is deep and spreading work increases wait times, or excel
when speed heterogeneity allows better load balancing. The results here show why both topology and hardware
regime matter and provide a solid foundation for studying these more realistic scenarios. Understanding how learned
scheduling strategies depend on workflow topology and power-speed correlations is the first step toward building robust,
topology-aware and regime-aware systems that can handle the full complexity of production cloud environments.

\nocite{*}
\bibliographystyle{ACM-Reference-Format}
\bibliography{references}

\end{document}